\documentclass[extendedabs]{recpad2k}
\usepackage{url}
\usepackage{hyperref}

\title{Evaluating Model Performance with Hard-Swish Activation Function Adjustments}

\addauthor{Sai Abhinav Pydimarry, Shekhar Madhav Khairnar, Sofia Garces Palacios, Ganesh Sankaranarayanan, Huu Phong Nguyen}{huuphong.nguyen@utsouthwestern.edu}{1}
\addauthor{Darian Hoagland, Dmitry Nepomnayshy}{}{2}
\addinstitution{
 Department of Surgery, University of Texas Southwestern Medical Center, TX, USA
}
\addinstitution{
Lahey Hospital and Medical Center, Burlington, MA, USA
}

\runninghead{Student, Prof, Collaborator}{RECPAD Author Guidelines}

\begin{document}

\maketitle

\begin{abstract}
In the field of pattern recognition, achieving high accuracy is essential. While training a model to recognize different complex images, it is vital to fine-tune the model to achieve the highest accuracy possible. One strategy for fine-tuning a model involves changing its activation function. Most pre-trained models use ReLU as their default activation function, but switching to a different activation function like Hard-Swish could be beneficial. This study evaluates the performance of models using ReLU, Swish and Hard-Swish activation functions across diverse image datasets. Our results show a $2.06\%$ increase in accuracy for models on the CIFAR-10 dataset and a 0.30\% increase in accuracy for models on the ATLAS dataset. Modifying the activation functions in architecture of pre-trained models lead to improved overall accuracy.
\end{abstract}

\section{Introduction}
\label{sec:intro}
Neural networks were first introduced in the mid of last century, and the term ``deep learning'' (DL) gained prominence in the mid-2000s due to the work of Geoffrey Hinton and others, which led to a tremendous amount of research in this area. At the heart of all deep learning networks are activation functions, which add non-linearity to neurons, enabling them to solve more complex problems. Rectified Linear Units (ReLUs) have been widely adopted into many DL models due to its simplicity and effectiveness. Recently, a newer activation function, Swish, demonstrated that it could match or outperform ReLU, increasing the overall performance of models~\cite{ramachandran2017searching}. However, Swish is computationally intensive compared to ReLU. To address this, a state-of-the-art activation function called Hard-swish was introduced. It is a piece-wise function that uses ReLU6 and can provides a very close approximation to Swish, preserving most of the benefits of improved accuracy and performance while making the computations faster~\cite{howard2019searching}. In this work, we initially examine the performance of the Hard-Swish activation function on the standard CIFAR-10 image dataset for image classification using the ResNet18 model, before evaluating its effectiveness on our ATLAS dataset with the X3D model for phase segmentation.

\section{Activation Functions}
\label{sec:proposed}
\subsection{ReLU}
Rectified Linear Units (ReLU) is one if the most commonly used activation functions for neural networks $ReLU(x) = \max(0, x)$. 


\subsection{Sigmoid}
Like ReLU, the Sigmoid activation function is used in neural networks with the formula : $Sigmoid(x) = \frac{1}{1 + e^{-x}}$.
\subsection{Swish}
The Swish activation function is a newer function that builds on the previous two activation functions mentioned (it was discovered by using automatic search techniques to find new activation functions)~\cite{ramachandran2017searching} $Swish(x) = x \cdot Sigmoid(x) = x \cdot \frac{1}{1 + e^{-x}}$. 
%
%
\subsection{Hard-Swish}
The Hard-Swish activation function is a piece-wise approximation of the Swish activation function~\cite{howard2019searching}. It retains the positive aspects of the Swish activation function while reducing the computational resources $Hard-Swish(x) = x \cdot \frac{ReLU6(x + 3)}{6}$ and $ReLU6(x) = \min(\max(0, x), 6)$.
%
\begin{figure} [tbh!]
\begin{center}
\includegraphics[keepaspectratio,width=0.45\textwidth]{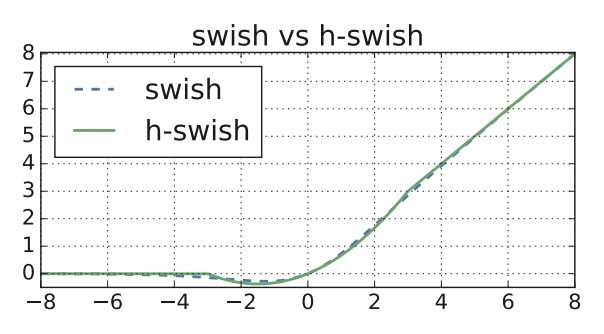}
\caption{Hard-Swish vs Swish~\cite{howard2019searching}}
\label{fig:HardSwish_Diagram}
\end{center}
\end{figure}
\section{Materials and Methods}
\label{sec:method}
\subsection{ResNet}
The ResNet18 Model from the PyTorch library was used for the CIFAR-10 dataset (See Section \ref{sec:CIFAR-10}). ResNet18 is a deep convolutional neural network, known for its using residual blocks with identity skip connections. The architecture begins with a convolution and max-pooling layer, followed by four groups of residual blocks, and ends with a fully connected layer. Residual connections enable the model to learn from the difference between the input and output, rather than learning a traditional mapping from input to output. Each residual block in the model architecture uses a ReLU activation function, which we modified for model evaluation. 

\subsection{X3D}
The X3D-m model from PyTorch was used for ATLAS dataset (See Section~\ref{sec:atlas}). It was pre-trained on the Kinetics-400 dataset~\cite{kay2017kinetics}, which includes 400 human action classifications. We then trained and fine-tuned it to recognize distinct phases from the ATLAS. The X3D-m model consists of six main layers, including the initial layer, ResStages 1 through 4, and a global max pooling layer. Each ResStage contains residual blocks, with the numbers varying from stage to stage. Stage one has three blocks, Stage two has five blocks, Stage three has eleven blocks, and Stage four has seven blocks. These blocks have two activation functions: act\_a (Band A), set to ReLU, and act\_b (Band B), set to Swish~\cite{feichtenhofer2020x3d}. The model was modified to replace its activation functions from the default ones to the Hard-Swish. Each experiment was run twice, and the mean results were computed. The second and third ResStages were combined into the ``Middle Layers''. The default model's accuracy served as the baseline with no changes to its activation function. In post-processing, we implemented the Smooth Moving Average (SMA)~\cite{guinon2007moving} with the window size $w$ varying.

\section{Experiments}
\label{sec:experiments}
\subsection{Benchmark Dataset}
\subsubsection{CIFAR-10}
\label{sec:CIFAR-10}
CIFAR-10 is a standard benchmark dataset for DL, which comprises $60000$ colored images, size $32 \times 32$ pixels, and has 10 labels, each with $6000$ images. When preparing the data for training, we used CIRFAR10’s predetermined sets for training and testing from PyTorch. The training dataset has a total of $50000$ images, and the testing dataset has $10000$ images. 

\subsubsection{ATLAS}
\label{sec:atlas}
\begin{figure} [tbh!]
\begin{center}
\includegraphics[keepaspectratio,width=0.45\textwidth]{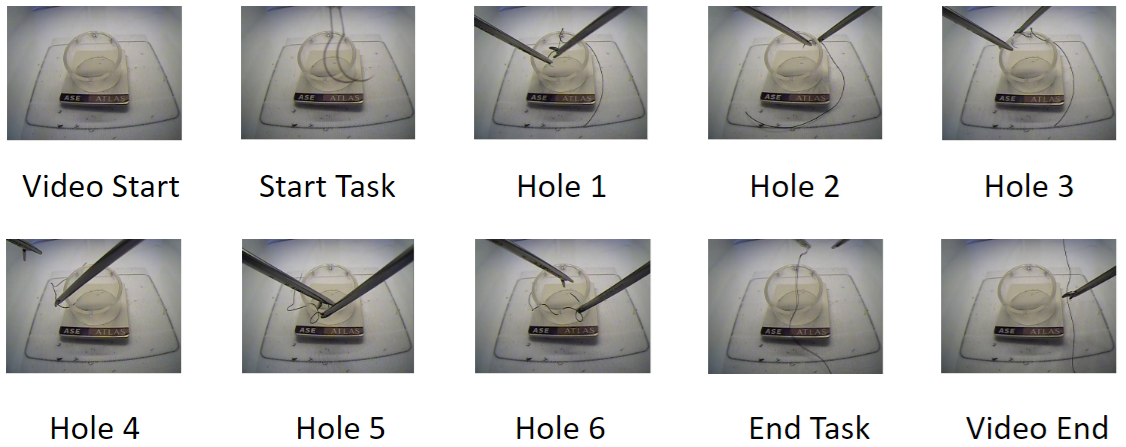}
\caption{Ten Phases in ATLAS Task 1.}
\label{fig:task1phases}
\end{center}
\end{figure}

The Advanced Training in Laparoscopic Suturing (ATLAS) program, designed by the Association for Surgical Education (ASE) in 2022, comprises Task 1 for which 49 videos were collected. Out of these videos, 39 were used for training ($32042$ images), with the remaining 10 for testing ($10209$ images). 
The task is split into ten distinct phases, including Video Start, Task Start, Hole 1-6, Task End, and Video Stop (Figure~\ref{fig:task1phases}).
%
%
%
\section{Classification Results}
\label{sec:result}
\subsection{Performance - CIFAR-10}
 Overall, performance varied depending on the layers where the activation functions were modified. In the ResNet18 models that were trained on CIFAR-10 dataset, the model with the initial layer’s activation function changed from ReLU to Hard-Swish achieved the highest average accuracy of $83.81\%$, outperforming the baseline by $2.06\%$. We observed that the training time was also varied: the model that had all of the activation functions changed to Hard-Swish took the longest time of 1004 seconds. The model with the best accuracy -- the initial layer -- took 624 seconds.
\begin{figure} [hbt!]
\begin{center}
\includegraphics[keepaspectratio,width=0.45\textwidth]{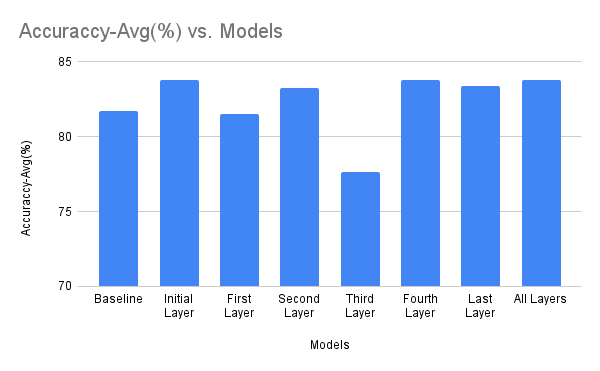}
\caption{CIFAR-10\ Results - Accuracy}
\label{fig:cifar}
\end{center}
\end{figure}
%

\subsection{Performance act\_a - ATLAS}
\label{subsec:act_a}
\begin{figure} [hbt!]
\begin{center}
\includegraphics[keepaspectratio,width=0.45\textwidth]{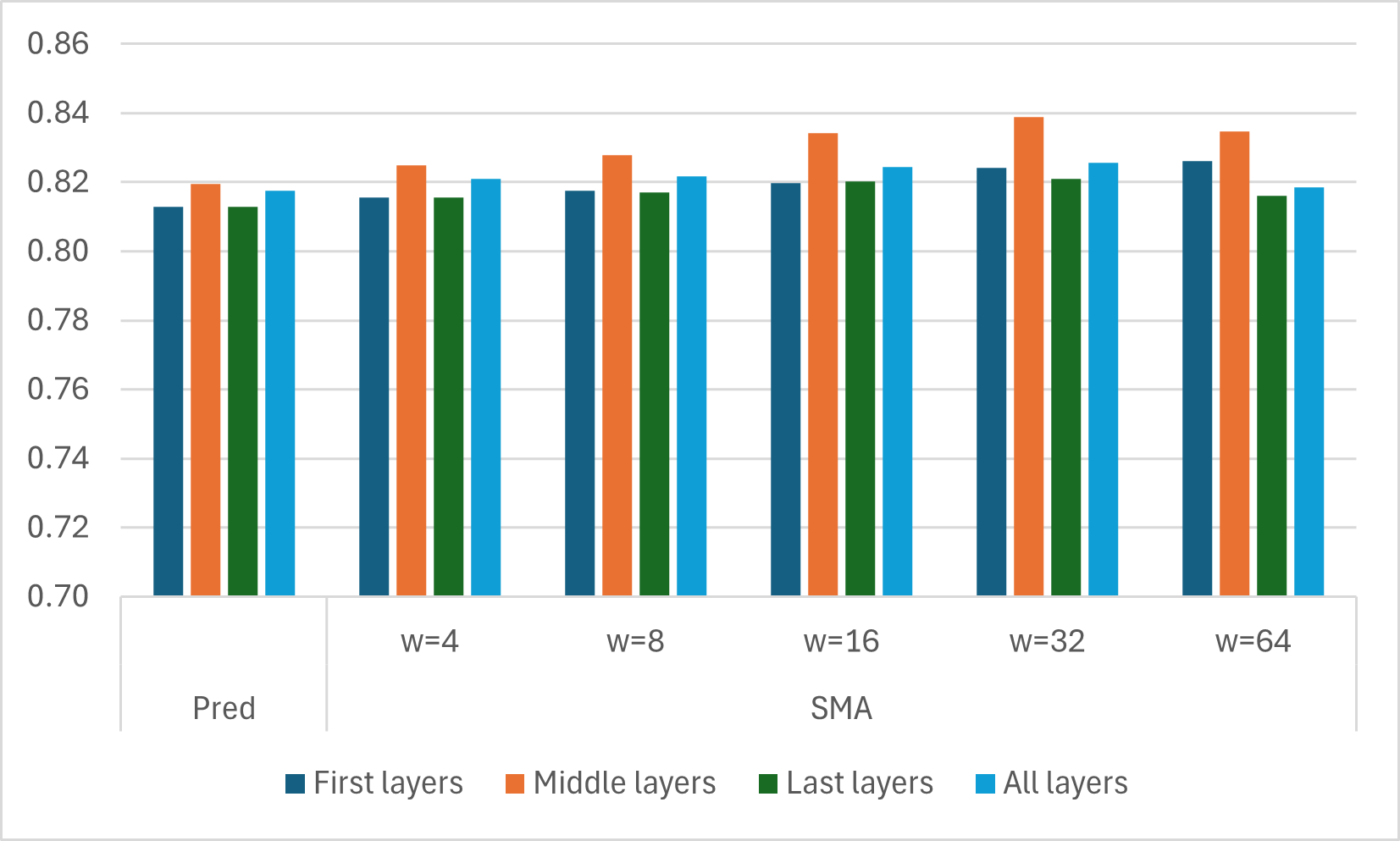}
\caption{ATLAS Phase Segmentation Accuracy. Changed ReLU Activation Function to Hard-Swish on Band A.}
\label{fig:hswish_a}
\end{center}
\end{figure}

In this experiment, the X3D-m model was trained on the ATLAS Dataset. The highest accuracy of the baseline was $83.57\%$ when SMA was used with the window size $w$ was set to $32$. On the other hand, the highest average accuracy was $83.88\%$ (increased $0.31\%$) when all blocks of act\_a were switched from ReLU to Hard-Swish in the Middle Layers and $w$ was set to $32$ (Figure~\ref{fig:hswish_a}). 
\subsection{Performance act\_b - ATLAS}
\begin{figure} [hbt!]
\begin{center}
\includegraphics[keepaspectratio,width=0.45\textwidth]{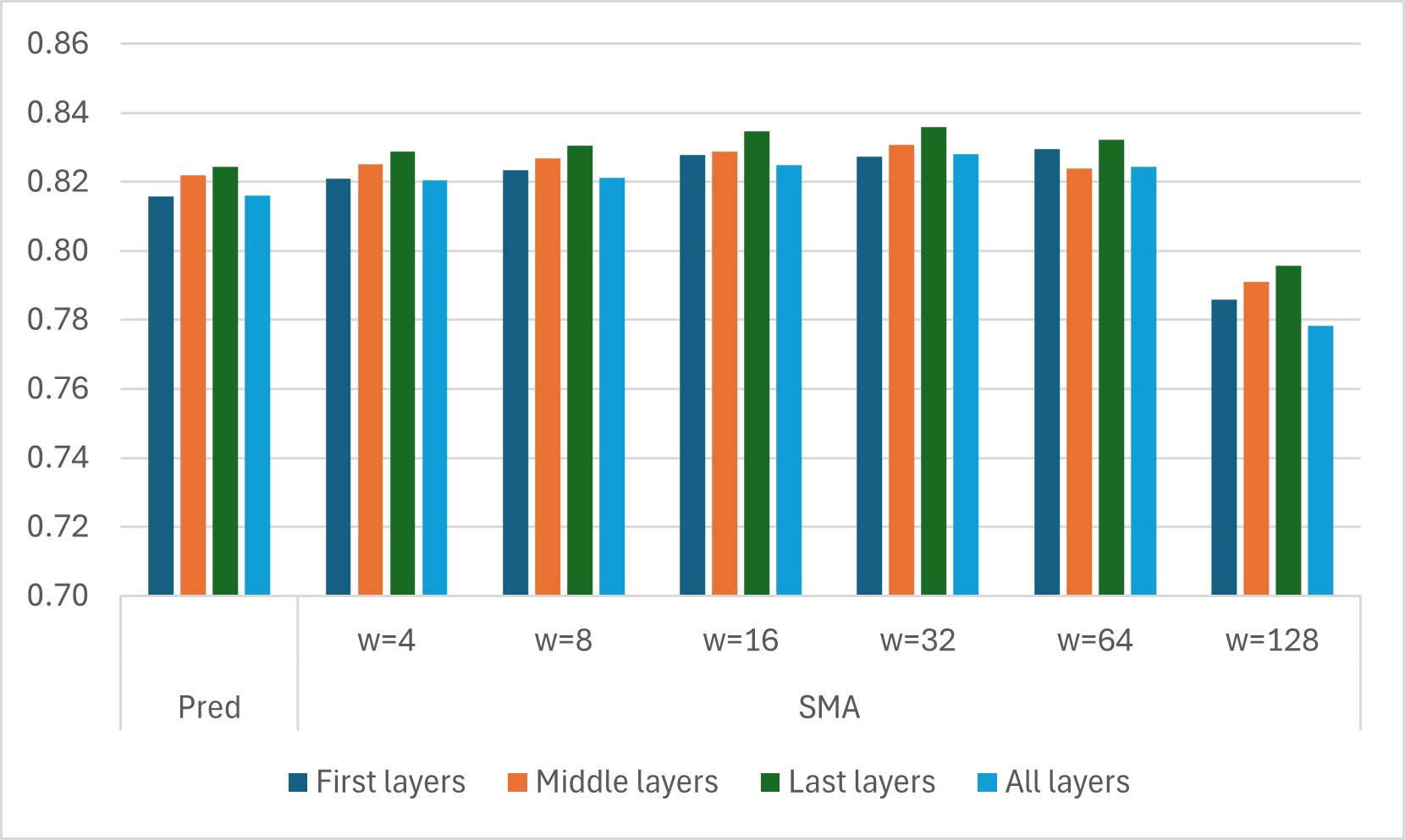}
\caption{ATLAS Phase Segmentation Accuracy. Changed Swish Activation Function to Hard-Swish on Band B.}
\label{fig:hswish_b}
\end{center}
\end{figure}

In a similar manner, we repeated the previous analysis with the activation functions of the blocks act\_b modified from Swish to Hard-Swish. As we can see from the Figure~\ref{fig:hswish_b}, overall performance improved as the window size 
$w$ increased, but declined when $w$ was increased beyond 64. The highest average accuracy was $83.58\%$ when all instances of act\_b were changed from Swish to Hard-Swish in the Last Layer and $w$ was set to 32. Though the performance is comparable, Hard-Swish is more efficient as it requires less computation power than the Swish.

%
\section{Conclusion}
\label{sec:conclusion}
Activation functions are crucial to neural networks, and some offering significant advantages over others. Modifying activation functions in pre-trained models such as X3D-m for phase segmentation on the ATLAS dataset and ResNet18 for image classification on CIFAR-10 can improve both overall accuracy and efficiency.
\section*{Acknowledgments}
We thank Lahey Hospital Medical Center for granting access to the ATLAS dataset.

\bibliography{aknn}

\begin{thebibliography}{5}
\providecommand{\natexlab}[1]{#1}
\providecommand{\url}[1]{\texttt{#1}}
\expandafter\ifx\csname urlstyle\endcsname\relax
  \providecommand{\doi}[1]{doi: #1}\else
  \providecommand{\doi}{doi: \begingroup \urlstyle{rm}\Url}\fi

\bibitem[Feichtenhofer(2020)]{feichtenhofer2020x3d}
Christoph Feichtenhofer.
\newblock X3d: Expanding architectures for efficient video recognition.
\newblock In \emph{Proceedings of the IEEE/CVF conference on computer vision and pattern recognition}, pages 203--213, 2020.

\bibitem[Gui{\~n}{\'o}n et~al.(2007)Gui{\~n}{\'o}n, Ortega, Garc{\'\i}a-Ant{\'o}n, and P{\'e}rez-Herranz]{guinon2007moving}
Jos{\'e}~Luis Gui{\~n}{\'o}n, Emma Ortega, Jos{\'e} Garc{\'\i}a-Ant{\'o}n, and Valent{\'\i}n P{\'e}rez-Herranz.
\newblock Moving average and savitzki-golay smoothing filters using mathcad.
\newblock \emph{Papers ICEE}, 2007:\penalty0 1--4, 2007.

\bibitem[Howard et~al.(2019)Howard, Sandler, Chu, Chen, Chen, Tan, Wang, Zhu, Pang, Vasudevan, et~al.]{howard2019searching}
Andrew Howard, Mark Sandler, Grace Chu, Liang-Chieh Chen, Bo~Chen, Mingxing Tan, Weijun Wang, Yukun Zhu, Ruoming Pang, Vijay Vasudevan, et~al.
\newblock Searching for mobilenetv3.
\newblock In \emph{Proceedings of the IEEE/CVF international conference on computer vision}, pages 1314--1324, 2019.

\bibitem[Kay et~al.(2017)Kay, Carreira, Simonyan, Zhang, Hillier, Vijayanarasimhan, Viola, Green, Back, Natsev, et~al.]{kay2017kinetics}
Will Kay, Joao Carreira, Karen Simonyan, Brian Zhang, Chloe Hillier, Sudheendra Vijayanarasimhan, Fabio Viola, Tim Green, Trevor Back, Paul Natsev, et~al.
\newblock The kinetics human action video dataset.
\newblock \emph{arXiv preprint arXiv:1705.06950}, 2017.

\bibitem[Ramachandran et~al.(2018)Ramachandran, Zoph, and Le]{ramachandran2017searching}
Prajit Ramachandran, Barret Zoph, and Quoc~V Le.
\newblock Searching for activation functions.
\newblock \emph{ArXiv}, abs/1710.05941, 2018.
\newblock URL \url{https://api.semanticscholar.org/CorpusID:10919244}.

\end{thebibliography}
\end{document}